\documentclass{article}

\usepackage{microtype}
\usepackage{graphicx}
\usepackage{subfigure}
\usepackage{booktabs} 

\usepackage{hyperref}
\usepackage{enumitem}
\usepackage{booktabs}
\usepackage{siunitx}
\usepackage{tabularx}
\usepackage{multirow}
\usepackage{adjustbox}
\usepackage{color}

\usepackage[accepted]{icml2025}

\usepackage{amsmath}
\usepackage{amssymb}
\usepackage{mathtools}
\usepackage{amsthm}
\usepackage{todonotes}
\usepackage{xcolor}
\usepackage[capitalize]{cleveref}

\theoremstyle{plain}

\theoremstyle{definition}

\theoremstyle{remark}

\icmltitlerunning{On Fairness of Unified Multimodal Large Language Model for Image Generation}

\begin{document}

\twocolumn[
\icmltitle{On Fairness of Unified Multimodal Large Language Model for Image Generation}



\icmlsetsymbol{equal}{*}

\begin{icmlauthorlist}
\icmlauthor{Ming Liu}{i}
\icmlauthor{Hao Chen}{c}
\icmlauthor{Jindong Wang}{w}
\icmlauthor{Liwen Wang}{i}
\icmlauthor{Bhiksha Raj Ramakrishnan}{c}
\icmlauthor{Wensheng Zhang}{i}

\end{icmlauthorlist}

\icmlaffiliation{i}{Iowa State University}
\icmlaffiliation{c}{Carnegie Mellon University}
\icmlaffiliation{w}{William Mary University}


\icmlkeywords{Machine Learning, ICML}

\vskip 0.3in
]



\printAffiliationsAndNotice{}  

\begin{abstract}
Unified multimodal large language models (U-MLLMs) have demonstrated impressive performance in \emph{visual understanding} and \emph{generation} in an end-to-end pipeline. 
Compared with \emph{generation-only} models (e.g., Stable Diffusion), U-MLLMs may raise new questions about \emph{bias} in their outputs, which can be affected by their unified capabilities. This gap is particularly concerning given the under-explored risk of propagating harmful stereotypes.
In this paper, we benchmark the latest U-MLLMs and find that most of them exhibit significant demographic biases, such as gender and race bias. 
To better understand and mitigate this issue, we propose a \emph{locate-then-fix} strategy, where we \emph{audit} show how the individual model component is affected by bias.
Our analysis shows that bias originates primarily from the language model. 
More interestingly, we observe a ``partial alignment'' phenomenon in U-MLLMs, where understanding bias appears minimal, but generation bias remains substantial.
Thus, we propose a novel \emph{balanced preference loss} to balance the demographic distribution with synthetic data.
Experiments demonstrate that our approach reduces demographic bias while preserving semantic fidelity. 
We hope our finding underscores the need for more holistic interpretation and debiasing strategies of U-MLLMs 
in the future. 
\end{abstract}

\vspace{-3em}
\section{Introduction}
\label{sec:intro}
 
\begin{figure}[ht]
    \centering
    \includegraphics[width=0.47\textwidth]{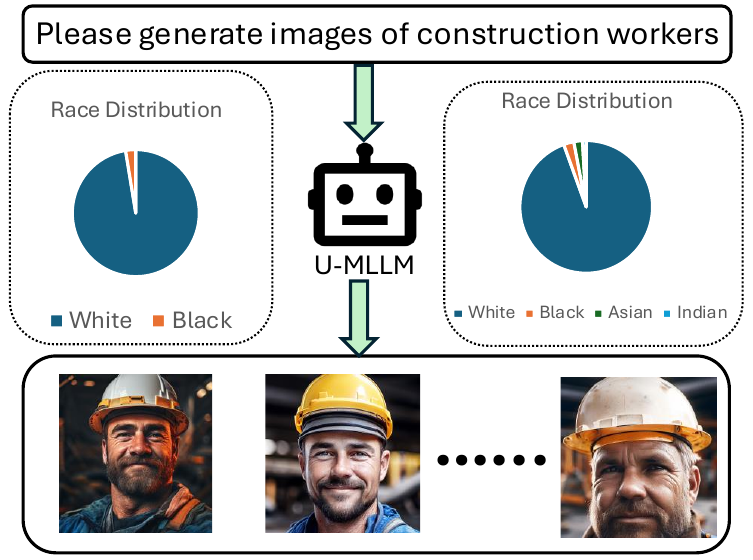}
    \vspace{-0.5cm}
    \caption{U-MLLMs are capable of generating images with high quality, but the generation lacks diversity. The model has a bias for some occupations. In this example, given the prompt ``construction worker'', the model generate most of images with demographic attribute as ``male'' and ``white''.}
    \label{fig:fairness-mllms}
    \vspace{-0.75cm}
\end{figure}

Multimodal Large language models (MLLMs) have demonstrated remarkable capabilities in visual understanding \cite{li2024llava, wang2024qwen2vlenhancingvisionlanguagemodels}. Recent research \cite{wu2024vila,  wu2024janus} has focused on extending MLLMs' capabilities to image generation settings, enabling them to produce textual content and \emph{visual} content. These \emph{unified} MLLMs (U-MLLMs), e.g., VILA-U \cite{wu2024vila}, present both visual understanding and generation capability, where they can not only understand the semantics in images but also generate images with high-quality conditioning on user prompts in natural language. However, these U-MLLMs with unified capabilities may inadvertently reproduce or amplify \emph{biases} at deployment, including gender and racial stereotypes embedded in their large-scale training data \cite{elazar2023s,chen2024catastrophic}. 

A common structure in U-MLLMs is an image \emph{tokenizer} that transforms images into discrete tokens via an \emph{encoder-decoder} framework \cite{esser2021taming}. Specifically, the vision encoder compresses the input image into latent embeddings and then quantizes them into discrete tokens. Afterward, a decoder reconstructs the image from these tokens. This discrete tokenization bridges textual and visual modalities by analogizing image tokens to words in a sentence, enabling a single autoregressive objective that unifies text and image generation. While this design has proven effective for quality and scalability, it also opens additional channels for bias from the tokenizer.

Existing work on debiasing in image generation has highlighted the social risks posed by skewed output distributions, and various methods have been proposed to reduce bias in image generation \cite{chuang2023debiasingvisionlanguagemodelsbiased,bansal2022texttoimagegenerativemodelsunderstand,wang2024conceptalgebrascorebasedtextcontrolled,gandikota2024unifiedconcepteditingdiffusion}. 
Nevertheless, many such methods are designed specifically for diffusion models, which leverage different principles for image generation \cite{shen2024finetuningtexttoimagediffusionmodels, gandikota2024unifiedconcepteditingdiffusion}. 
As U-MLLMs with autoregressive image generation capabilities become more and more prevalent, it is imperative to evaluate their bias in image generation and develop new methods to reduce bias in these token-based generation models. Moreover, it remains an open question whether the generation biases emerge more from the vision encoder, which generates image tokens, or from the language modeling component, which generates image tokens according to the given text prompt.

This paper investigates and mitigates demographic bias in U-MLLMs for \emph{text-to-image} generation. Specifically, we made the first step to study gender and race bias for U-MLLMs. We benchmark the latest U-MLLM on gender and race bias using datasets and metrics introduced by a recent study of image generation fairness \cite{shen2024finetuningtexttoimagediffusionmodels}. 
These models comprehensively include VILA-U \cite{wu2024vila}, Show-o \cite{xie2024showo}, Janus \cite{wu2024janus}, Janus-Pro \cite{chen2025januspro}, TokenFlow \cite{qu2024tokenflow}, Emu3 \cite{wang2024emu3}. 
Our results show that the latest UMLLMs exhibit notable gender and race biases in image generation (see an example in \autoref{fig:fairness-mllms}). Next, we conduct a detailed audit of the vision encoder/decoder and the language model component to localize these biases' source(s), where we find that the biases are mainly from the language model. Finally, we synthesize high-quality training data with a balanced demographic distribution and propose a novel balance preference loss to mitigate generation bias in U-MLLMs, inspired by recent research on direct preference optimization \cite{rafailov2024direct, hong2024orpomonolithicpreferenceoptimization}.

Through extensive experiments with various U-MLLMs, we demonstrate that our approach significantly reduces biases (e.g., over-generation of certain genders or races) without sacrificing the quality of image generation output. For example, for the VILA-U model, our method reduces its gender bias by 71.91\% and increases the inception score by 12.2\%. In summary, our key contributions are:
\vspace{-1em}
\begin{itemize}[leftmargin=*]
    \item \textbf{Benchmarking Bias}: We benchmark the latest U-MLLMs on race and gender bias and find that they often display it with different degrees for image generation. Notably, Janus-Pro, the latest U-MLLM, witnessed the worst gender bias with a value of 0.90, compared with the stable diffusion model with a bias value of 0.67.
    \item \textbf{Localizing Bias}: We inspect different components in the VILA-U model (vision encoder, language model) by using methods such as linear probing in image embedding space to pinpoint potential sources of bias and find that the bias is mostly from the language model component.
    \item  \textbf{Mitigating Bias}: We used the diffusion model to synthesize training data with a balanced demographic distribution. We also introduce a balanced preference loss inspired by direct preference optimization, with the objective of balancing the likelihood of visual generation towards the different demographic groups. We empirically show that our approach yields substantial improvement in demographic fairness while preserving the quality of image generation, thus providing a practical framework for developing unified MLLMs with more fairness.
\end{itemize}

\vspace{-2em}
\section{Preliminary}
\label{sec:pre}

\paragraph{Structure of U-MLLMs}
We consider an \emph{autoregressive} U-MLLM that, given a textual prompt $x$, first converts it into a sequence of text tokens $\{x_1, \dots, x_{T_x}\}$ and then generates a sequence of \emph{image tokens} $\{z_1,\dots,z_{T_z}\}$, which an image decoder reconstructs into a final image $y$ (see~\autoref{fig:unified_model} for the pipeline). Let $\boldsymbol{\theta} = \{\boldsymbol{\theta}_{\mathrm{v}}, \boldsymbol{\theta}_{\mathrm{l}}\}$ denote the model parameters, where: \emph{(1)} $\boldsymbol{\theta}_{\mathrm{v}}$ is the \textbf{image tokenizer} (encoder-decoder) responsible for converting input images into discrete tokens and decoding tokens back into images. \emph{(2)}$ \boldsymbol{\theta}_{\mathrm{l}}$ is the \textbf{language model} (LM) that processes and generates token sequences (both text and image tokens) in a unified autoregressive way.

As shown in \autoref{fig:vision_tower},  the image encoder $E_{\boldsymbol{\theta}_{\mathrm{v}}}$ maps an image $y$ into latent embeddings $\textbf{e}$, then quantizes the embeddings into a discrete token sequence $\{z_1,\dots,z_{T_z}\}$. Conversely, the image decoder $D_{\boldsymbol{\theta}_{\mathrm{v}}}$ inverts this process:
\begin{align}
    \{z_1,\dots,z_{T_z}\} &= E_{\boldsymbol{\theta}_{\mathrm{v}}}(y), 
    \label{eq:enc}\\
    y &= D_{\boldsymbol{\theta}_{\mathrm{v}}}(z_1,\dots,z_{T_z}).
    \label{eq:dec}
\end{align}
Meanwhile, the LM $LM_{\boldsymbol{\theta}_{\mathrm{l}}}$ treats both text tokens and image tokens uniformly under a single next-token probability:
\begin{equation}
    P_{\boldsymbol{\theta}}(z_t \mid x,\, z_{<t})
    \;=\;
    LM_{\boldsymbol{\theta}_{\mathrm{l}}}(z_{t-1},\dots,z_1;\, x).
    \label{eq:lm}
\end{equation}
This design allows the U-MLLM to perform \emph{visual understanding} by mapping an image to a semantic space (via \cref{eq:enc}) and then interpreting those discrete tokens as inputs to the LM, and to perform \emph{image generation} by autoregressively sampling $\{z_t\}$ from \cref{eq:lm} and reconstructing an image from the resulting token sequence via \cref{eq:dec}.
\vspace{-1em}
\paragraph{Demographic Bias.}  
When conditioning on neutral generation prompt (no explicit gender or race specified), the model could exhibit \emph{demographic bias} by disproportionately generating output with a distribution skewing towards certain demographic groups (e.g., “male” or “female,” “Asian” or “Black”). ~\autoref{fig:fairness-mllms} shows such an example where a prompt \emph{``Please generate images of construction workers''} yield mostly images samples of one demographic group.

Formally, let $d \in \mathcal{D}$ where $D = \{d_1, \ldots, d_K\}$ is a set of demographic labels (e.g., \{male, female\} or \{Asian, Black, White, Indian\}). Given a neutral prompt \( x \), such as ``a portrait of a construction worker'', an unbiased model would generate a list of images \( \{y_1, y_2, \ldots\} \) with these demographic labels in a balanced distribution, for example, 50-50 for gender, or a uniform distribution for race. By contrast, we find that the latest U-MLLMs generate
\[
P_\theta(\mathcal{C}(y) = d_i \mid x) \gg P_\theta(\mathcal{C}(y) = d_j \mid x),
\]
where $d_i, d_j$ corresponds to ``male'',``female'' respectively(in this case) and $\mathcal{C}$ is a pre-trained image classifier from prior study \cite{shen2024finetuningtexttoimagediffusionmodels} that labels each image $y$ with a predicted attribute $\hat{d}_i$. This indicated a strong bias in gender preferences. Our goal is to mitigate this bias while preserving overall image fidelity.
\vspace{-1em}
\paragraph{Direct Preference Optimization} As an alternative to computational expensive RLHF \cite{ouyang2022traininglanguagemodelsfollow} methods that are required to train a reward model separately, DPO re-parameterizes reward model $r$ via a ratio between the policy $\pi_\theta$ and a reference policy $\pi_{\text{ref}}$ and eliminates the cost of training a reward function $r(x,y)$ explicitly \cite{rafailov2024direct}. Concretely, for a given pair $(x,y)$,
\begin{align}
r(x,y) 
&= \beta \log \frac{\pi_\theta(y \mid x)}{\pi_{\text{ref}}(y \mid x)} 
+ \beta \log Z(x),
\label{eq:dpo_reward_restate}
\end{align}
where $Z(x)$ is a partition function and $\beta$ is a scaling factor\cite{rafailov2024direct}. By plugging $r$ into a Bradley-Terry model\cite{Bradley1952RankAO}, DPO formulates the preference probability for a winning response $y_w$ over a losing response $y_l$ as
\begin{equation}
p(y_w \succ y_l \mid x) 
\;=\;
\sigma \bigl( r(x,y_w) - r(x,y_l)\bigr),
\label{eq:bt-model}
\end{equation}
DPO aims to optimize the policy \emph{directly} by minimizing \autoref{eq:bt-model} and increasing the difference in models' preferences between $y_w$ and $y_l$ without a separate reward model.

To further reduce the overhead of the reference model, recent work explores \emph{reference-free} preference optimization~\cite{hong2024orpomonolithicpreferenceoptimization, meng2024simposimplepreferenceoptimization}. One representative method, ORPO~\cite{hong2024orpomonolithicpreferenceoptimization}, defines the \emph{odds} of generating a response $y$ given prompt $x$ as
\begin{equation}
    \text{odds}_\theta(y \mid x) 
    \;=\; 
    \frac{p_\theta(y \mid x)}{1 - p_\theta(y \mid x)}
\end{equation}

and further quantifies the odds ratio between two responses $y_{w}$ and $y_{l}$ via
\begin{equation}
    \text{OR}_\theta\bigl(y_{w}, y_{l}\bigr)
    \;=\;
    \frac{\text{odds}_\theta(y_{w} \mid x)}{\text{odds}_\theta(y_{l} \mid x)}.
\end{equation}
and finally, define a preference loss term as
\begin{equation}
    \mathcal{L}_{OR} = -\log \sigma \left( \log \frac{odds_\theta(y_w|x)}{odds_\theta(y_l|x)} \right) \label{eq:ratio} 
\end{equation}
By encouraging a large odds ratio $\text{OR}_\theta(y_{w}, y_{l})$, the model is pushed to \emph{prefer} the response $y_{w}$ over $y_{l}$ \emph{directly}, without relying on a separate  “reference” model. 
\vspace{-1em}
\section{Locating Bias}
As shown in~\autoref{fig:locating_bias}, to identify the biases that might emerge from \emph{where}, we analyze intermediate outputs - that is, the \emph{image tokens} (produced by the LM) and the \emph{image embeddings} (produced by the vision encoder) during the decoding step. We consider two main hypotheses regarding the origin of demographic bias as follows.
\vspace{-1em}
\subsection{Hypothesis I: The Bias is from Vision Encoder}
The vision encoder transforms input images into a high-dimensional embedding space, as illustrated in~\autoref{fig:vision_tower}. This encoder itself may be a source of bias within U-MLLMs. To test this hypothesis, we focus on auditing the \emph{vision encoder}, which is trained symmetrically alongside the decoder. By detecting the embedding output from the encoder, we aim to identify any inherent biases that could influence the overall fairness of the model.

\vspace{-1em}
\paragraph{Linear Probing on Image Embeddings}
\autoref{fig:locating_bias} illustrated the overall pipeline to audit vision encoder: first, we sample a balanced set of images denoted as ${y_i}$ from FairFace\cite{kärkkäinen2019fairfacefaceattributedataset} covering different demographic attributes (e.g., male, female). By feeding these images into the vision encoder, we obtained a set of image embeddings denoted as $\mathbf{E} = \{\mathbf{e}_i\}$. Each embedding $\mathbf{e_i}$ is labeled with the ground truth attribute $d_i \in \{ \text{male}, \text{female} \}$ according to its corresponding image input $y_i$. We then split $\mathbf{E}$ into training and testing subsets and train a \emph{linear} classifier $\ell(\cdot)$ to predict demographic labels from embeddings $e_i$. This gives us pairs $\bigl(y_i, \hat{d}_i, \mathbf{e}_i\bigr)$—i.e.\ the image, the predicted demographic label, and the corresponding image embedding. The precision, F1 score, recall and precision in the test set are all \emph{high}, indicating that the encoder’s latent space preserves explicit demographic distinctions:
\[
\widehat{d}_i \;=\; \ell(e_i) \quad \rightarrow \quad \mathrm{High\ Accuracy}.
\]
Since the decoder is trained to invert the encoder, it should \emph{faithfully} preserve these attributes. We also find that when we prompt the model with an attribute (e.g., ``Asian professor''), the final generated images are overwhelmingly associated with that attribute. This also implies that the \emph{decoder} consistently reflects whatever demographic semantics are either present in the tokens or passed by the encoder. Therefore, the chance that bias originated from the vision encoder is small. 
\vspace{-1em}
\subsection{Hypothesis II: The bias is from the language model}
Let $x_{\text{neutral}}$ be a neutral prompt (e.g., ``a photo of a professor''), and let $x_{\text{aug}}(d_i)$ denote an \emph{augmented} version of the same prompt specifying a particular demographic $d_i$ (e.g., ``a photo of a female professor''). Assume that for each $x$, we sample a list of images $\{y_1, \ldots, y_M\}$ and each image $y_i$ is associated with a sequence of image tokens generated from LM as $\mathbf{z}_i = (z_{i,1}, z_{i,2}, \ldots)$.

\vspace{0.5em}
\paragraph{Collecting Image Tokens.}
We record the \emph{intermediate} image tokens for (\emph{1}) The neutral prompt $x_{\text{neutral}}$  and (\emph{2}) The demographic-augmented prompts $x_{\text{aug}}(d_i)$ as described above. We then use a pre-trained image classifier $\mathcal{C}$ \cite{shen2024finetuningtexttoimagediffusionmodels} to label each final image $y_i$ with a predicted attribute $ \widehat{d}_i \;=\; \mathcal{C}(y_i)$. This yields pairs $\bigl(x_i, y_i, \hat{d}_i, \mathbf{z}_i\bigr)$—i.e.\ the prompt, the resulting image, the predicted demographic label, and the corresponding image token sequence.
\vspace{-1em}
\paragraph{Distribution Over Image Tokens.}
Let $p_{\theta}(z \mid x)$ denote the probability distribution over the image tokens $z$ given a prompt $x$. In practice, we approximate it by a finite sample:
\[
\widehat{p}_{\theta}(z \mid x)
\;=\;
\frac{1}{M}\,\sum_{i=1}^{M} \delta(\mathbf{z}_i),
\]
where $\delta(\cdot)$ is a Kronecker delta that counts the sampled sequences $\{\mathbf{z}_i\}$. This empirical approximation estimates the token distribution \( p_{\theta}(z \mid x) \) by averaging over \( M \) generated image token sequences, effectively capturing the model's likelihood of producing each token \( z \) given the prompt \( x \).

We then calculate a \emph{distribution distance}, e.g., Jensen-Shannon Divergence(JSD)\cite{Lin1991DivergenceMB}, between the neutral-prompt distribution $\widehat{p}_{\theta}(z \mid x_{\text{neutral}})$ and the augmented-prompt distribution $\widehat{p}_{\theta}(z \mid x_{\text{aug}}(d_i))$. Formally,
\[
D_{\mathrm{JS}}\!\Bigl(
\widehat{p}_{\theta}(z \mid x_{\text{neutral}})\;\|\;
\widehat{p}_{\theta}(z \mid x_{\text{aug}}(d_i))
\Bigr).
\]
where
\[
D_{\mathrm{JS}}(P\|Q) = \frac{1}{2}D_{\mathrm{KL}}(P\|M) + \frac{1}{2}D_{\mathrm{KL}}(Q\|M)
\]
and $M = \frac{1}{2}(P + Q)$ is the average distribution, with $D_{\mathrm{KL}}$ denoting the Kullback-Leibler divergence.

We observe that when $d_i$ matches the \emph{predicted demographic label} $\hat{d}_i$ predicted from the final images, the JSD between the two distributions is significantly smaller. In other words, the token-level distribution for the neutral prompt $x_{\text{neutral}}$ is \emph{closest} to that for $x_{\text{aug}}(d_i)$ corresponding to the majority demographic group the model actually generated. 
\vspace{-1em}
\paragraph{Takeaway} This indicates that the \emph{language model} itself already introduces demographic preferences \emph{at the token level}. If the prompt ``firefighter'' leads to a token distribution similar to ``male firefighter,'' it suggests that the tokens $\mathbf{z}_i$ are implicitly biased toward that demographic, even \emph{before} any visual decoding procedure. Hence, the major source of demographic bias remains in the upstream processes: the autoregressive image token generation. In the following section, we describe the approach we developed to mitigate the debias in image generation by focusing on \emph{LM}.

\vspace{-1em}
\section{Method}
\label{sec:method}

To effectively reduce bias in U-MLLM, a training dataset that \emph{explicitly} includes diverse demographic attributes is of importance. In general, web-scale data under-represents certain groups or inherently skews toward certain stereotypes. By synthesizing balanced data—where each demographic group appears in similar proportion- we can give the model a stronger signal to reduce its biases during training. This can ensure that the U-MLLM is trained on a \emph{controlled distribution} counteracts the bias present in real-world data distribution.

\subsection{Training Data Generation}
\label{subsection: training data generation}

We leveraged diffusion model FLUX.1-dev\cite{flux2023} to synthesize our training data as follows:
\vspace{-1em}
\begin{enumerate}[leftmargin=1em]
    \item \textbf{Base Prompt Selection.}  
    We begin by collecting a set of \emph{base prompts} $x_{neutral}$ (for example, ``a portrait of a \{occupation\}''), where occupations are drawn from a publicly available dataset\cite{shen2024finetuningtexttoimagediffusionmodels}. 
    
    \item \textbf{Demographic Augmentation.}  
    For each base prompt $x_{neutral}$, we augment it with demographic attributes $d \in \{d_1, d_2, \dots\}$, resulting in \emph{augmented prompts} $x_{\text{aug}}(d_i)$. For example, starting with the base prompt ``a photo of a nurse'' we generate augmented prompts such as ``a photo of a male nurse'' and ``a photo of an Asian nurse.''  
    
    \item \textbf{Image Generation via Diffusion Model.}  
    For each $x_{\text{aug}}(d_i)$, we feed it into FLUX.1-dev\cite{flux2023} model to obatin a set of images $\{y_1, y_2, \dots\}$. This step ensures diversity in the visual representations of each demographic group.
    
    \item \textbf{Paired Dataset Creation.}  
    To create the dataset, we pair each generated set of images $y_k$ with its corresponding base prompt $x_{neutral}$. Before pairing, we use the ChatGPT-4o model to paraphrase the base prompt to introduce linguistic diversity while maintaining neutrality. For instance, the base prompt ``a photo of a nurse'' can be paraphrased as ``a photo of an individual in the nursing profession.'' The resulting synthetic dataset pairs each neutral prompt with a set of images, where each image corresponds to a specific demographic group:
    \[
        \mathcal{D}_{\mathrm{syn}} 
        \;=\; \bigl\{\,(x_{\mathrm{neutral}},\, y_{d_1},..., y_{d_K})\,\bigr\}.
    \]
\end{enumerate}

Because we explicitly injected demographic attributes into the prompts, $\mathcal{D}_{\mathrm{syn}}$ spans multiple genders, ethnicities, and roles—offering more balanced coverage than typical real-world data.

\subsection{Balanced Preference Loss}

\begin{figure*}[ht]
    \centering
    \vspace{-0.25cm}
    \includegraphics[width=0.8\textwidth]{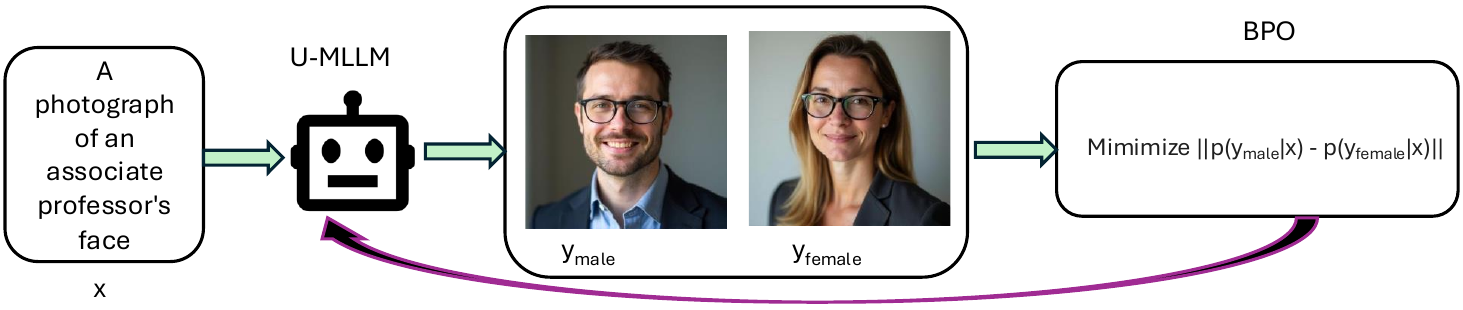}
    \vspace{-0.5cm}
    \caption{The optimization objective is to minimize the deviation of preference between different demographic group.}
    \vspace{-0.5cm}
    \label{fig:bpo-loss}
\end{figure*}



Existing preference-optimization approaches focus on \emph{aligning} the model’s response to \emph{user-specified} preferences (e.g., “prefer polite over rude text”). By contrast, in this study we seek \emph{balance} among demographic attributes---for instance, the model should equally generate \{female, male\} or \{Asian, Black, Indian, White\} under neutral prompts, rather than overwhelmingly generate one demographic group. This shift in perspective means we want the \textbf{absolute difference in preference} across demographics to be \emph{minimized}, rather than pushing a single demographic to be more or less likely in isolation. To encourage equal representation, as shown in \autoref{fig:bpo-loss}, we aim to penalize deviations by minimizing
\begin{equation}
\min ||p(y_{d_i} \succ y_{d_j} \mid x) -
p(y_{d_j} \succ y_{d_i} \mid x) ||
\end{equation}

In other words, if the model is more likely to generate $y_{d_1}$ (image with demographic label $d_1$) than $y_{d_j}$ (image with demographic label $d_j$), it will be penalized, which encourages balance of the output distribution, and vice versa.

\paragraph{Definition} Concretely, let $y_{d_i}, y_{d_j}$ be images from different demographic groups. We define a \emph{balanced odds} penalty:
\begin{equation}
  \mathcal{L}_{\mathrm{bal}}(\theta)
  \;=\;
  \log\left[ 1 + \left( 
    \sigma\!\Biggl(
      \log\text{OR}_\theta(y_{d_i}, y_{d_j}) 
    \Biggr) - \frac{1}{2} \right)^2 \right]
  \label{eq:bal_twogroups}
\end{equation}

Minimizing \eqref{eq:bal_twogroups} can lead to $p(y_{d_i} \succ y_{d_j} \mid x)$ closer to $p(y_{d_j} \succ y_{d_i} \mid x)$, thus encouraging the even distribution of generation among demographic attributes. 

\paragraph{Extension to Multiple Groups.}  
For race attributes, we have multiple demographic groups. Let's define $K$ to be the number of race categories. The auxiliary loss to penalize demographic odds of different attributes deviating from each other:
\begin{equation}
  \mathcal{L}_{\mathrm{bal}}(\theta)
  \;=\;
  \sum_{k=1}^{K}
  \mathcal{L}_{\mathrm{bal}}^{(d_{k},d_{(k+1) \% K})}(\theta).
  \label{eq:multigroup}
\end{equation}

\paragraph{Two-Stage Training} We adopt a straightforward two-stage procedure as illustrated in Algorithm  \ref{alg:balanced_finetuning}.
\begin{enumerate}[leftmargin=1em]
    \item \textbf{Supervised Finetuning (SFT).} 
    We begin by finetuning the U-MLLM on a supervised dataset of \{\emph{prompt}, \emph{image tokens}\} pairs to ensure high-quality image generation and semantic fidelity. This yields an SFT model $\pi_{\text{SFT}}$.
    \item \textbf{Balanced Preference Optimization.} 
    We then apply a reference-free odds-ratio penalty via Eq. \ref{eq:multigroup} to make the model equally prefer multiple demographic attributes.   
\end{enumerate}

Our approach \emph{directly} integrates demographic balance into the policy’s objective via~\autoref{eq:multigroup}. By including samples with multiple demographic attributes in the balanced dataset $\mathcal{D}_{\text{bal}}$, the model is enabled to generalize towards each demographic group. The balanced preference loss $\mathcal{L}_{\mathrm{bal}}$ then reduces the inter-difference among demographic groups for the model's preference. Overall, this two-stage method ensures the model maintains good generation quality (from Stage 1) while significantly reducing \emph{absolute preference} differences between demographic attributes (from Stage 2).

\vspace{-1em}
\section{Experiment}

\subsection{Experimental Setup}
\label{subsec:exp_setup}



\paragraph{Models.}
In our study, we evaluate the generation bias with respect to gender and race for latest U-MLLMs that can both understand and generate visual content. The models we considered include VILA-U\cite{wu2024vila}, TokenFlow\cite{qu2024tokenflow}, Emu3\cite{wang2024emu3}, Janus\cite{wu2024janus}, Show-o\cite{xie2024showo}, Janus-Pro\cite{chen2025januspro}. Since VILA-U was the only model with available training code when we began this study, we fine-tuned it into various variants for comparison. For VILA-U, we compare its variants: \textbf{\emph{(1). }Prompt engineering}: original VILA-U model with a modified prompt to explicitly ask for diversity in image generation. \textbf{\emph{(2). }Understanding Finetuning(\textsf{I~$\rightarrow$~T})}: VILA-U finetuned on the dataset in \emph{image-to-text} fashion.  \textbf{\emph{(3). }Generation Finetuning(\textsf{T~$\rightarrow$~I})}: VILA-U finetuned on the same balanced data in \emph{text-to-image} fashion.\textbf{\emph{(4). }Balanced Preference Loss}: VILA-U finetuned using our proposed balanced preference optimization for visual generation. (Sec. \ref{sec:method}).
\vspace{-1.5em}
\paragraph{Evaluation Data.}
For evaluation, we collect a set of occupation-based generation prompts (e.g. ``nurse,'' ``data engineer,'' ``senator'') that have been used in previous studies\cite{shen2024finetuningtexttoimagediffusionmodels} to assess diffusion models' biases concerning gender and race. These prompts are 50 in size and are publicly available. We utilized these prompts to evaluate various U-MLLMs under consistent conditions: We generate $N = 160$ images for each test prompt and use an attribute classifier\cite{shen2024finetuningtexttoimagediffusionmodels} to determine the demographic attribute labels.
\vspace{-1.5em}

\paragraph{Finetuning Data.}
For training, we collected a separate set of occupation-based generation prompts\cite{shen2024finetuningtexttoimagediffusionmodels} (e.g., ``scientist,'' ``firefighter'') and generated \emph{synthetic} images using a diffusion model, as described in Section~\ref{sec:method}. Our dataset comprises 1,000 training prompts, resulting in 10,000 images per demographic group. We consider six demographic groups: male, female, White, Black, Asian, and Indian. Our training dataset consists of 10,000 samples, each containing a text prompt and six images from different demographic groups.
 
\textbf{Metrics and Evaluation} We follow prior study\cite{shen2024finetuningtexttoimagediffusionmodels}, the same \textbf{bias metrics} and \textbf{image quality metrics} are adapted:
\begin{itemize}[leftmargin=*]
    \vspace{-1em}
    \item \textbf{Bias Metrics}: Representation Disparity (RD) 
    measures how often each demographic attribute appears in generated outputs given a neutral prompt (e.g., ``CEO''). We leverage the classifier model provided by prior study to predict the attributes from generated images. RD is calculated as:
    \[
            \mathrm{bias}(\mathbf{P}) 
            \;=\; 
            \frac{1}{K(K-1)/2} \sum_{i,j \in [K] : i < j} \bigl|\mathrm{freq}(i) - \mathrm{freq}(j)\bigr|,
    \]
    where $\mathrm{freq}(i)$ is the frequency of demographic $d_i$ in the image generation samples. A lower RD indicates less biased and more balanced demographic representation.
    \vspace{-1em}
    \item \textbf{Image Quality Metrics}: we utilized a range of metrics for image quality.  \textbf{\emph{(1)}. CLIP Score}
    Measures alignment between generated image and textual prompt. Higher scores indicate better semantic matching.
\textbf{\emph{(2)}. CLIP IQA}:
    Image quality assessment using CLIP's visual-textual understanding. Evaluates aspects such as sharpness without reference images.
 \textbf{\emph{(3)}. Inception Score}:
    Metric used for evaluating quality and diversity of generated images. Higher scores indicate better quality.
\end{itemize}
\vspace{-1em}
\paragraph{Evaluation Protocol.}
First, we generate $N = 160$ images for each test prompt. Next, we apply a demographic classifier from a previous study \cite{shen2024finetuningtexttoimagediffusionmodels} to predict the labels of demographic attributes for each image. These labeled attributes are then used to compute the overall bias and semantic score based on the above metrics. 
\vspace{-1em}
\paragraph{Training Procedure.} \emph{Base Learning Rate}: We start with a fixed learning rate (for example, $1\times10^{-4}$) for fine-tuning. \emph{Batch Size}: Ranges from 8 to 32, depending on the setup. \emph{Number of Steps}: For the first stage, we fine-tune for up to 10 epochs,  checking bias and quality after the training.  For the second stage, the epoch is chosen to be 1 to 2. We use the LoRA\cite{hu2021loralowrankadaptationlarge} method for finetuning; the rank is 32 for all experiments. 



\vspace{-1em}
\subsection{Experimental Results}
\label{sec:exp_results}

\newcommand{\graysmall}[1]{{\color{gray}\scriptsize #1}}
\newcommand{\STAB}[1]{\begin{tabular}{@{}c@{}}#1\end{tabular}}
{
\begin{table*}[ht]
    \setlength{\tabcolsep}{8pt}
    \centering
    \small
    \vspace{-0.5cm}
    \caption{Image generation bias.}
    \label{table:gender_race_bias}
    \adjustbox{max width=1.0\textwidth}{%
    \begin{tabular}{c | r | c c c c c c c}
        \toprule
        \multirow{2}{*}{\STAB{\rotatebox[origin=c]{90}{Debias:~}}} & \multirow{2}{*}{Method} & \multicolumn{3}{c}{Bias $\downarrow$} & \multicolumn{3}{c}{Semantics Preservation $\uparrow$} \\
        \cmidrule(lr){3-5}
        \cmidrule(lr){6-8}
         & & Gender & Race & G.$\times$R. & CLIP-S  & CLIP-IQA & Inception \\
        \cmidrule(r){2-8}
        & Stable Diffusion & 0.67  & 0.42  & 0.21  & ---   & ---   & ---   \\
        & Janus           & 0.87  & 0.43  & 0.23  & 27.44 & 0.69  & 2.27  \\
         & Janus-Pro           & 0.90  & 0.48  & 0.24  & 27.62 & 0.82  & 1.79  \\
        & Show-o          & 0.85  & 0.48  & 0.24  & 27.16 & 0.86  & 1.79  \\
        & TokenFlow       & 0.84  & 0.47  & 0.24  & 27.17 & 0.84  & 2.34  \\
        & Emu3            & 0.83  & 0.42  & 0.22  & 27.93 & 0.89  & 2.29  \\
        & VILA-U          & 0.89  & 0.48  & 0.24  & 28.24 & 0.84  & 1.87  \\
        \hline
        \multirow{4}{*}{\STAB{\rotatebox[origin=c]{90}{Gender}}}
        & Prompt Engineering      & 0.56 & 0.49 & 0.23 & 28.51  & 0.82 & 1.91 \\
        & Finetune(\textsf{I~$\rightarrow$~T}) & 0.83 & 0.42 & 0.22 & 28.49 & 0.83 & 2.28  \\
        & Finetune(\textsf{T~$\rightarrow$~I})  & 0.27 & 0.51 & 0.23 & 27.66 & 0.77 & 1.85 \\
        & BPO & \textbf{0.25} & 0.50 & 0.22 & 27.74 & 0.77 & 2.10                 \\
        \hline
        \multirow{5}{*}{\STAB{\rotatebox[origin=c]{90}{Race}}}
        & Prompt Engineering       & 0.56  & 0.49  & 0.23  & 28.51  &  0.82  &   1.91 \\
        & Finetune(\textsf{I~$\rightarrow$~T}) & 0.78  & 0.44  & 0.22  & 28.14  &  0.80 & 2.54    \\
        & Finetune(\textsf{T~$\rightarrow$~I})  & 0.83  & \textbf{0.23}  & 0.17  & 27.98  & 0.80  & 1.98 \\
        & BPO  &0.78 & 0.26 & 0.18 &  27.66 & 0.81   & 2.31&   \\
        \hline
        \multirow{5}{*}{\STAB{\rotatebox[origin=c]{90}{G.$\times$R.}}}
        & Prompt Engineering       & 0.59  & 0.33  & 0.18  & 28.09  &  0.80 & 1.87  \\
        & Finetune(\textsf{I~$\rightarrow$~T}) & 0.86 & 0.45 & 0.23 & 28.34 & 0.82 &2.22 \\
        & Finetune(\textsf{T~$\rightarrow$~I})  & 0.46  & 0.32  & 0.17  & 27.90  &  0.78 & 1.91  \\
        & BPO  & 0.52 & 0.26 & \textbf{0.15} & 27.78 & 0.80 & 2.06 \\
        \bottomrule
    \end{tabular}
    }
    \vspace{-.4cm}
\end{table*}
}

Table~\ref{table:gender_race_bias} summarizes the results for generation bias evaluation on a range of U-MLLMs and especially the VILA-U model with a few debiasing strategies applied. We measure \emph{gender} bias, \emph{race} bias, and their intersection (\emph{G.$\times$R.}), together with semantics-preservation metrics. Lower bias scores indicate more fairness, while higher semantics scores indicate more fidelity in the generation.
\vspace{-1em}
\paragraph{Baseline Models.} As shown in Table~\ref{table:gender_race_bias}, Stable Diffusion exhibits moderate gender bias (0.67) and race bias (0.42). Among U-MLLMs, Janus-series (Janus, Janus-Pro) show higher overall bias in gender (0.87–0.90), as well as high race and G×R scores (0.43–0.48 and 0.23–0.24). Show-o and TokenFlow both witness slightly lower bias. Compared with other U-MLLMs, Emu3 stands out with low gender (0.83) and G×R (0.22) biases, along with good semantic scores. Finally, VILA-U exhibits strong text–image alignment (CLIP-S of 28.24) but has high gender (0.89) and intersectional (0.24) bias, highlighting the need for reducing bias in U-MLLMs.
\vspace{-1em}
\paragraph{Debiasing Gender.} Table~\ref{table:gender_race_bias} presents four methods to reduce gender bias in VILA-U. 
\emph{Prompt Engineering} witness moderate bias reduction (Gender = 0.56) while preserving high semantic alignment (CLIP-S = 28.51). 
\emph{Finetune (I~$\rightarrow$~T)} achieved the smallest reduction in gender bias (0.83) even though it witnessed a little increase in semantics (CLIP-S = 28.49). 
By contrast, \emph{Finetune (T~$\rightarrow$~I)} drastically lowers the gender bias to 0.27. However, it came with the side effect that image quality is lower (CLIP-IQA = 0.77). 
Finally, \emph{BPO} (Balanced Preference Optimization) further lowers gender bias (0.25) while preserving reasonable visual fidelity (Inception = 2.10), indicating that our approach is effective for mitigating stereotypical associations without compromising generation quality for U-MLLMs.
\vspace{-1em}
\paragraph{Debiasing Race.}
Table~\ref{table:gender_race_bias} shows the results for mitigating race bias in the VILA-U. 
\emph{Prompt Engineering} achieves a moderate bias reduction (Race = 0.49) while retaining a high level of semantic alignment (CLIP-S = 28.51). 
\emph{Finetune (I~$\rightarrow$~T)} yields Race = 0.44 and slightly lower semantic scores (CLIP-S = 28.14). 
By contrast, \emph{Finetune (T~$\rightarrow$~I)} significantly reduces the race bias to 0.23, albeit with some drop in semantic fidelity (Inception = 1.98). \emph{BPO} also lower the race bias a lot(0.26), and achieved good semantic fidelity (Inception 2.31) in the meantime.
\vspace{-1em}
\paragraph{Debiasing Intersectional Attributes (G$\times$R).}
As shown under \emph{Debias: G$\times$R} in Table~\ref{table:gender_race_bias}, we evaluate methods targeting \emph{both} gender and race simultaneously. 
\emph{Prompt Engineering} achieves a moderately low intersectional bias (G$\times$R = 0.18) while preserving acceptable semantic fidelity (CLIP-S = 28.09). 
\emph{Finetune (I~$\rightarrow$~T)} reduces intersectional bias to 0.23, accompanied by strong semantic alignment (CLIP-S = 28.34), whereas 
\emph{Finetune (T~$\rightarrow$~I)} further lowers G$\times$R to 0.17 but at the cost of slightly reduced image quality (Inception = 1.91). \emph{BPO} achieved the lowest bias score in gender-race intersection as 0.15. It also achieved high semantic fidelity with an Inception score of 2.06.

\subsection{Discussion}
\begin{figure*}[ht]
    \centering
    \vspace{-0.2cm}
    \includegraphics[width=1.0\textwidth]{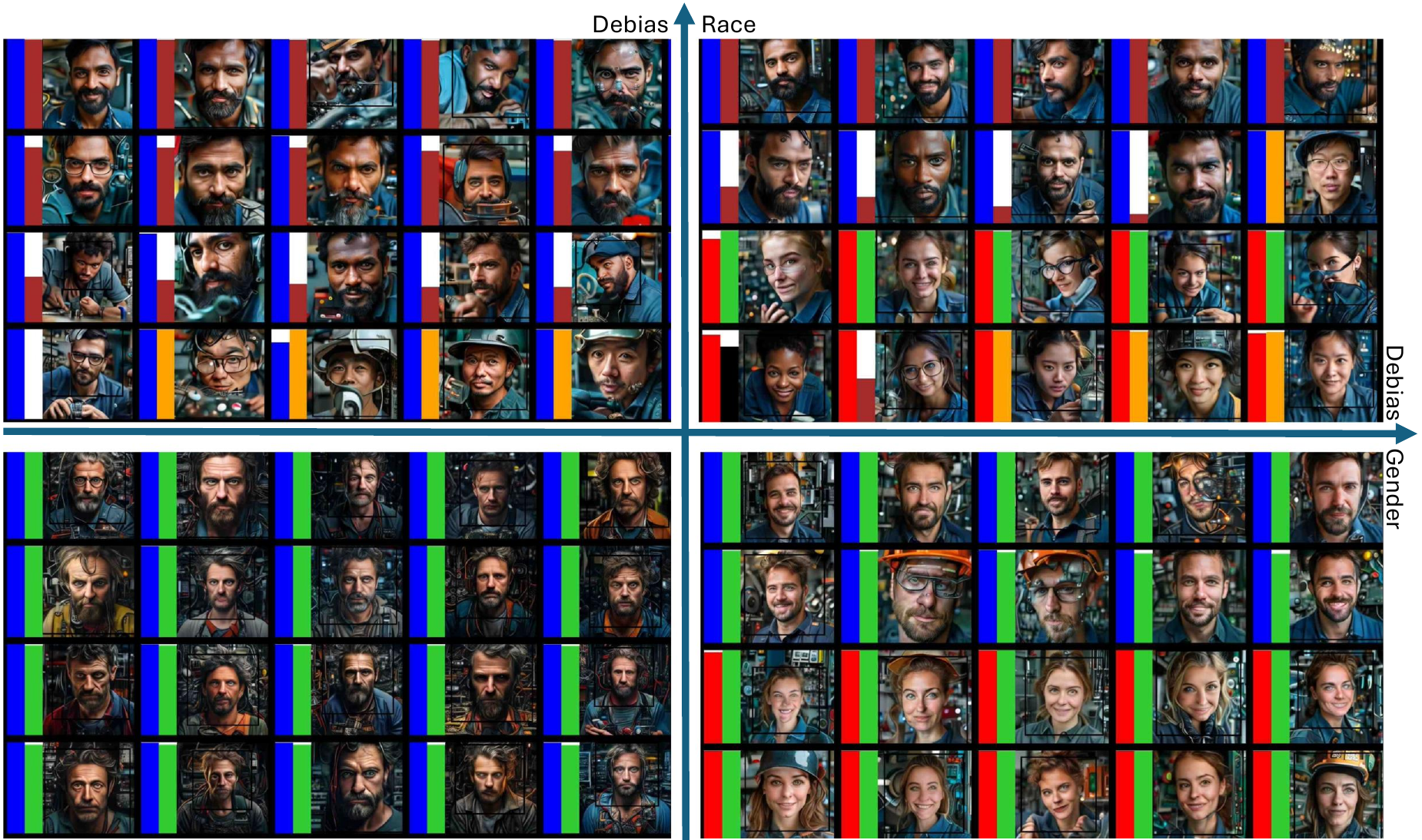}
    \vspace{-0.8cm}
    \caption{Given the test prompt: “a photo of the face of a electrical and electronics repairer, a person”. Images generated by: VILA-U (bottom-left), our method reduced bias in gender(bottom-right), race(top-left), gender-race(top-right). For each image, the first color-coded bar represents the predicted gender: \textcolor{blue}{male} or \textcolor{red}{female}. The second bar represents the predicted race: \textcolor{green}{While}, \textcolor{orange}{Asian}, Black, or \textcolor{brown}{Indian}. The height of each bar indicates the prediction confidence.}
    \label{fig:vilau-samples-compara}
    \vspace{-0.5cm}
\end{figure*}
\paragraph{Qualitive Samples}

In \autoref{fig:vilau-samples-compara}, we compare the VILA-U model with our method in different debiasing dimensions. Our approach evidently enables the model to generate more diverse samples. Our method effectively mitigates stereotypes and reduces the risk of perpetuating inequalities in automatically generated content.
\vspace{-1em}
\paragraph{Is there understanding bias in U-MLLMs?}
We examine \emph{visual understanding bias} to investigate bias in multimodal language models further. Specifically, we provide each U-MLLM with a prompt of the form \textit{``What is the gender of \{\texttt{occupation}\}? Answer with options male, female, unknown.''} while providing an input \emph{empty image} to ensure a fair comparison. We then sample 100 responses per prompt under a fixed sampling strategy. From these textual outputs, we extract the demographic attributes that each model most frequently associates with a given occupation. Then we use $RD$ metrics to calculate the bias.

\begin{figure}[ht]
    \centering
\includegraphics[width=0.48\textwidth]{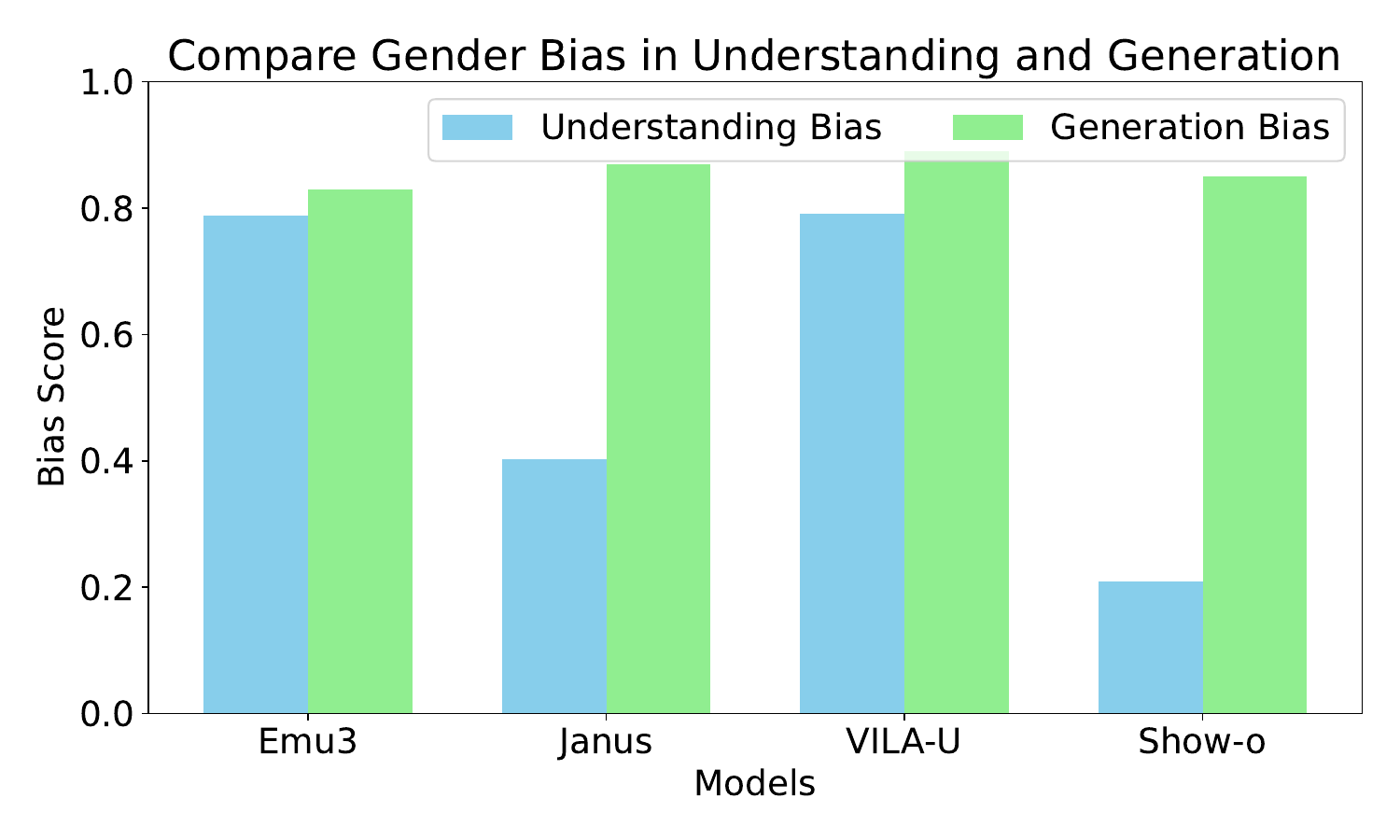}
    \vspace{-1cm}
    \caption{Compare the understanding and generation bias.}
    \vspace{-0.8cm}
    \label{fig:compare_bias}
\end{figure}

Results in Figure~\ref{fig:compare_bias} indicate that most models exhibit a pronounced \emph{understanding bias}; they systematically infer one gender or demographic for an occupation even when provided no actual visual cues. Interestingly, we observe cases of \emph{partial alignment}. For example, while the Janus model shows a moderate understanding bias in its responses---often refusing to label an individual’s gender or race explicitly, it nonetheless exhibits a substantial generation bias when prompted to create images. In other words, the model’s \emph{understanding-level} alignment (refusal to answer demographic queries) does not carry over to its \emph{image-generation} behavior, which remains biased. This highlights that merely aligning a model’s visual understanding to minimize bias in VQA scenarios is insufficient if its generative capability still systematically favors particular demographics. 

\paragraph{Can debiasing understanding help debias generation?}
Prior work\cite{tong2024metamorphmultimodalunderstandinggeneration} suggests that enhancing a model’s capability for image understanding can indirectly improve its performance in image generation. Motivated by this, we use the diffusion-generated images and their demographic labels (derived from prompts) to train the model in a \emph{discriminative} fashion. The resulting model fails to reduce demographic bias in \emph{image generation}. In other words, the ability to classify or reason about demographic attributes in images does not carry over to balanced \emph{generation} of images with varied demographics. 

These findings imply that bias in \emph{understanding} (e.g., naming attributes) and bias in \emph{generation} (e.g., synthesizing certain demographics disproportionately) may arise from distinct internal mechanisms. Whereas understanding tasks rely on discriminative representations, generative tasks involve autoregressive sampling or likelihood maximization that can perpetuate different stereotypes. Therefore, improving one aspect of visual-linguistic alignment does not necessarily resolve the other. Designing more advanced methods that holistically address both \emph{comprehension} and \emph{production} biases is essential for fair multimodal generation.

\vspace{-1em}
\paragraph{Generalization to other models.}

Table~\ref{tab:tokenflow_comparison} shows the results of applying similar finetuning strategies to \textbf{TokenFlow}. While TokenFlow initially exhibits a high gender bias (0.84), simple interventions such as \emph{prompting engineering} reduce its bias to 0.69.  \emph{Finetuning} approach further lowers it to 0.55, at a minor cost in semantics scores. These findings suggest that finetuning can be extended beyond VILA-U to other U-MLLM toward fairer image generation.



\section{Related Work}

\paragraph{Multimodal Generative Models}

Unified multimodal large language models (U-MLLMs) have advanced the start-of-the-art by bridging visual understanding and conditional generation capabilities. 
Compared to early MLLMs, which focus purely on understanding, such as Llava series \cite{liu2024visual,liu2023improvedllava}, these more recent works, represented by VILA-U \cite{wu2024vila}, Show-o~\cite{xie2024showo}, MetaMorph~\cite{tong2024metamorphmultimodalunderstandinggeneration}, TokenFlow~\cite{qu2024tokenflow}, Emu3~\cite{wang2024emu3}, TokenFusion \cite{zhou2024transfusion}, Janus \cite{wu2024janus,ma2024janusflow}, and etc. \cite{bachmann20244m,le2024diffusiongenerate,li2024dual}, highlight their effectiveness in generating high-quality visuals conditioned on text prompts.
U-MLLMs usually employ autoregressive paradigms that may inherit or amplify the biases embedded in their training data.
Existing studies predominantly focus on performance improvement rather than understanding and addressing the more critical issues such as demographic fairness.

\paragraph{Fairness in Image Generation}

The social risks of biased image generation, particularly gender and racial disparities, have been extensively documented \cite{kotek2023gender,li2024culturellm,li2024culturepark}. 
Prior efforts in diffusion-based models attempted to mitigate bias by re-balancing training data and incorporating fairness objectives \cite{kim2024training}. 
However, these approaches are not directly transferable to UMLLMs due to architectural differences and tokenization mechanisms. 
Few studies explore bias sources within unified models or their downstream effects on generated outputs, leaving a gap in understanding the interaction between text and image modalities.

\section{Conclusion}
\label{sec:limitation_conclusion}

\paragraph{Conclusion}
We examined demographic bias (e.g., gender, race) in unified multimodal large language models (U-MLLMs). We proposed a balanced preference optimization framework that combines supervised finetuning with a preference deviation penalty. Our experiments show that this method effectively reduces bias while maintaining high image quality and semantic fidelity. Moreover, our analysis of understanding versus generation bias underscores the importance of addressing \emph{both} visual reasoning and token-level preference shaping for comprehensive alignment. We hope these findings will guide future work toward building multimodal models with stronger fairness guarantees. Our work also have some limitation as illustrated in \autoref{sec:limitation}.

\paragraph{Impact Statement.}
Our work addresses the challenge of demographic bias in unified multimodal language models, aiming to ensure fair image generation for different demographic groups. Our methods can help mitigate harmful stereotypes and minimize the risk of reinforcing inequalities in automatically generated content. We believe ongoing work from researchers, policymakers, and the communities represented is important to maximize the societal benefits of debiasing techniques while maintaining transparency and fairness in machine learning systems.

\bibliography{example_paper}
\bibliographystyle{icml2025}

\newpage
\appendix
\onecolumn
\section{Structure of U-MLLM}
This section represents the structure of U-MLLM and visual tokenizer; all the content in this section is from prior study\cite{wu2024vila}. 

\begin{figure}[ht]
    \centering
    \includegraphics[width=0.9\textwidth]{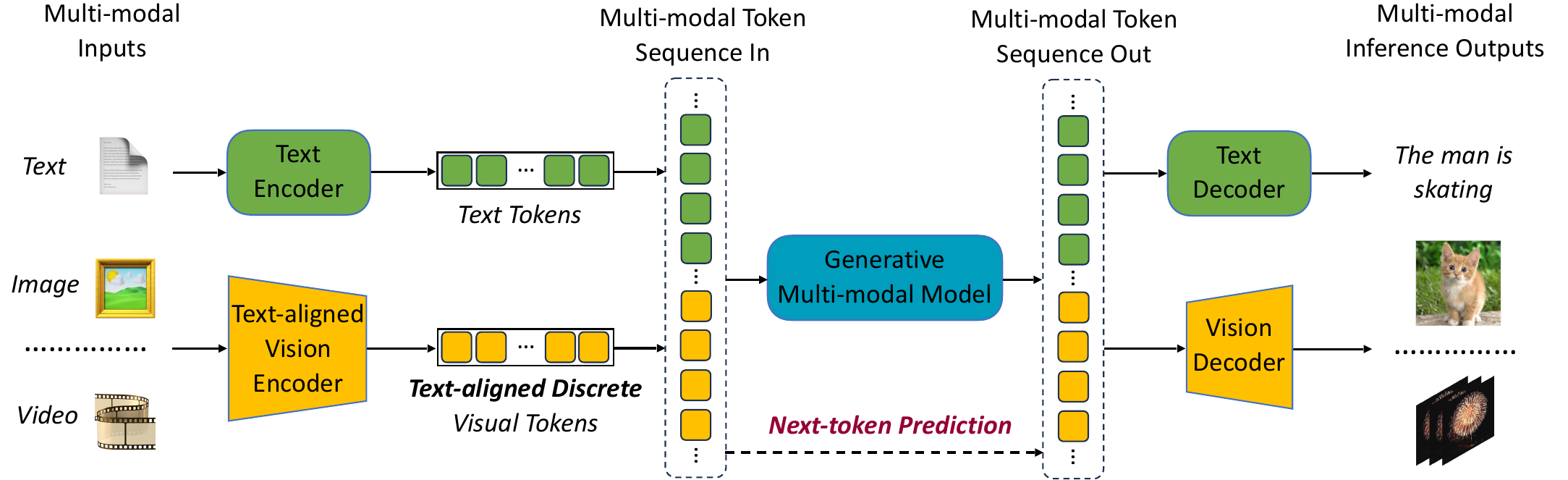}
    \caption{\textbf{Overview of framework’s multi-modal training and inference process\cite{wu2024vila}} Visual inputs are converted into discrete tokens and merged with textual tokens to create a unified multi-modal token sequence. This sequence is used in next-token prediction process, which supports a unified training objective. During inference, output tokens are processed through either text detokenizer or vision tower decoder, generating multi-modal content outputs\cite{wu2024vila}.}
    \label{fig:unified_model}
\end{figure}

\begin{figure}[ht]
    \centering
    \includegraphics[width=0.93\textwidth]{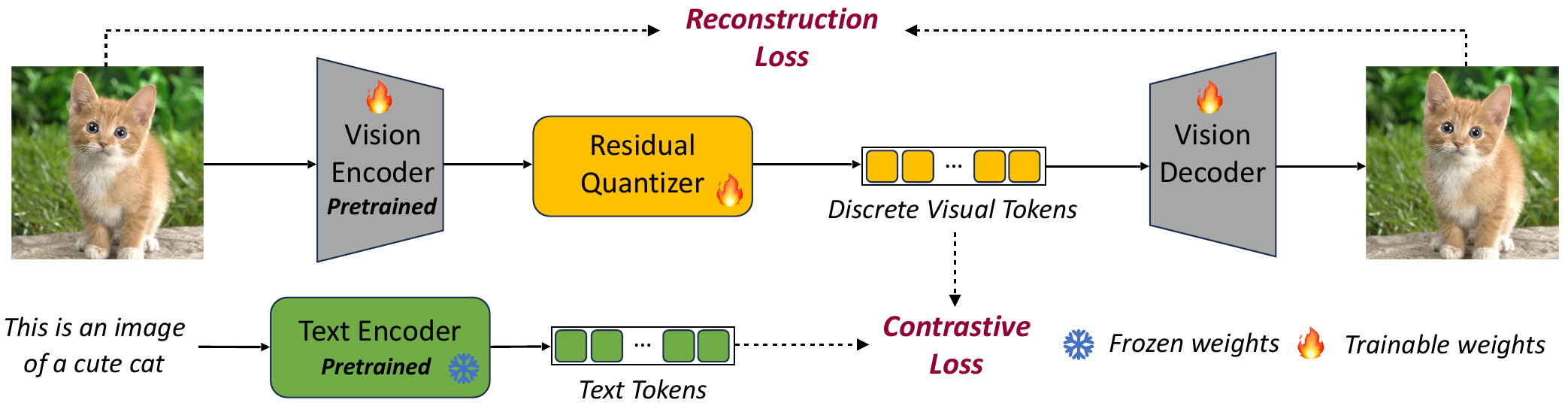}
    \caption{
    \looseness=-1
    \textbf{Overview of unified foundation vision tower\cite{wu2024vila}} Input images are processed by the vision encoder, where features are extracted and discretized using residual quantization. These discrete vision features are then utilized in two ways: they are fed into the vision decoder to reconstruct images and are used to perform text-image alignment. Throughout this process, both the reconstruction loss and contrastive loss are calculated to refine the vision tower, enabling it to generate discrete visual features that are aligned with text\cite{wu2024vila}.}
    \label{fig:vision_tower}
\end{figure}

\newpage
 
\section{Localizing Bias}
\label{subsec:localizing-bias}
\begin{figure*}[ht]
    \centering
    \vspace{-8pt}
    \includegraphics[width=0.93\textwidth]{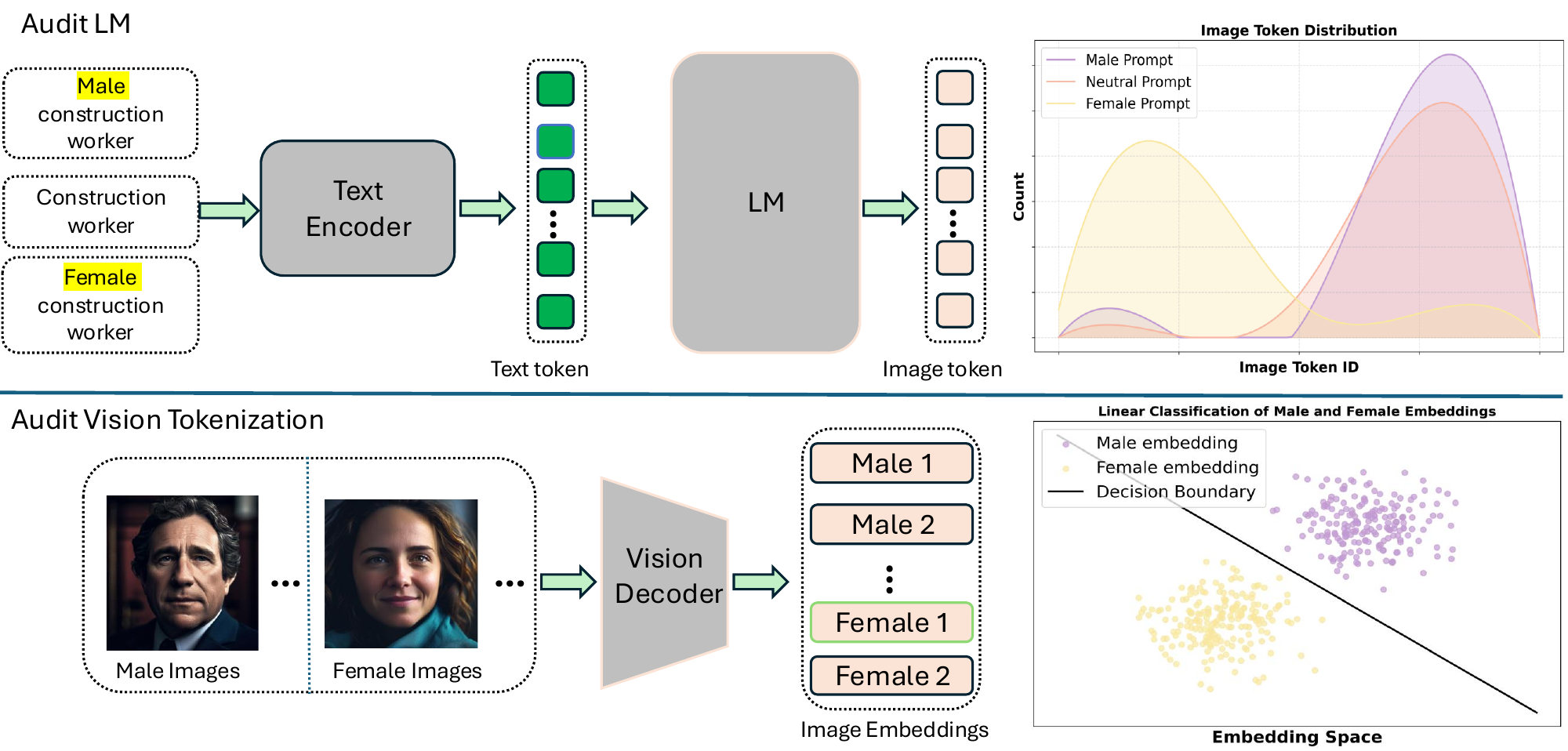}
    \caption{
    \looseness=-1
    \textbf{Detecting bias in LM(top), Vision encoder(bottom);}
    \label{fig:locating_bias}}
    \vspace{-10pt}
\end{figure*}

\section{BPO Algorithm}
\begin{algorithm}[ht]
\caption{Balanced Preference Optimization}
\label{alg:balanced_finetuning}
\textbf{Input}: U-MLLM with parameters $\theta_0$; SFT dataset $\mathcal{D}_{\text{SFT}} = \{(x_i, z_i)\}$ of prompts $x_i$ and image tokens $z_i$; Balanced dataset $\mathcal{D}_{\text{bal}} = \{(x_j, y_{j_1}, \ldots, y_{j_K})\}$, each $x_j$ with multiple demographic variants; Trade-off parameter $\lambda$, total training epochs $N_1, N_2$;
\textbf{Output}: Debiased model parameters $\theta$

\begin{algorithmic}[1]
\STATE \textbf{Stage 1: Supervised Finetuning}
\STATE \quad Initialize $\theta \leftarrow \theta_0$
\FOR{epoch $= 1$ to $N_1$}
    \STATE Sample a minibatch $\{(x_i, z_i)\}$ from $\mathcal{D}_{\text{SFT}}$
    \STATE $\mathcal{L}_{\mathrm{NLL}}(\theta)\;=\; -\sum_{(x_i,z_i)}\log P_\theta(z_i \mid x_i)$
    \STATE Update $\theta \leftarrow \theta - \eta \nabla_{\theta}\,\mathcal{L}_{\mathrm{NLL}}(\theta)$
\ENDFOR

\STATE \textbf{Stage 2: Balanced Preference Optimization}
\FOR{epoch $= 1$ to $N_2$}
    \STATE Sample a minibatch $\{(x_j, y_{j_1},\ldots,y_{j_K})\}$ from $\mathcal{D}_{\text{bal}}$
    \STATE \textbf{Balanced Preference Loss:} 
    \[
       \mathcal{L}_{\mathrm{bal}}(\theta)
       =
       \sum_{k\neq l}
        \mathcal{L}_{\mathrm{bal}}^{(d_{k-1},d_k)}(\theta).
    \]
    \STATE Update $\theta \leftarrow \theta - \eta \nabla_{\theta}\,\mathcal{L}_{\mathrm{bal}}(\theta)$
\ENDFOR
\STATE \textbf{Return} $\theta$
\end{algorithmic}
\end{algorithm}
\newpage
\section{Related Work(continued)}
\label{sec:related_work_continue}
\paragraph{Preference Optimization}  
Direct preference optimization (DPO) \cite{rafailov2024direct} has emerged as a promising technique to address biases in machine learning models, especially LLM. Since then, numerous new loss functions have been proposed~\cite{meng2024simposimplepreferenceoptimization, park2024disentanglinglengthqualitydirect, hong2024orpomonolithicpreferenceoptimization, ethayarajh2024ktomodelalignmentprospect, azar2023generaltheoreticalparadigmunderstand}.
Recent advances \cite{amini2024direct} integrate preference modeling into loss functions to guide models toward balanced outputs, which have rarely been explored for MLLMs.
Based on this, we introduce a novel balanced preference loss tailored for U-MLLMs.
By leveraging demographic attributes during training, the proposed method balances the likelihood of generating outputs across groups without compromising the image quality.

\section{Results for Debiasing TokenFlow}

\begin{table}[ht]
    \centering
    \small 
    \setlength{\tabcolsep}{2pt} 
    \vspace{-1.5em}
    \caption{Comparison of Gender Bias and Semantics Preservation Metrics for TokenFlow model.}
    \label{tab:tokenflow_comparison}
    \begin{tabular}{l|c|ccc}
        \toprule
        \textbf{Method} & \textbf{Gender Bias ↓} & \textbf{CLIP-S} & \textbf{CLIP-IQA} & \textbf{Inception} \\
        \midrule
        TokenFlow & 0.84 & 27.17 & 0.84 & 2.34 \\
        Prompting & 0.69 & 27.46 & 0.80 & 2.34 \\
        Finetune & 0.55 & 26.87 & 0.81 & 1.98 \\
        \bottomrule
    \end{tabular}
\end{table}

\section{Limitation}
\label{sec:limitation}
\paragraph{Limitation}
Several questions remain open despite the bias reduction achieved by our approach. First, our study primarily centered on \emph{overt} demographic categories (e.g., gender, race). Real-world scenarios may demand addressing \emph{intersectional} or \emph{nuanced} attributes (e.g., age, culture, or religion). Second, many models are not fully open-source, restricting the scope of our evaluations to two publicly available systems. Future research could broaden the range of tested models. Lastly, due to resource constraints, we did not explore alternative preference optimization objectives beyond our current framework. Building on our method to incorporate other debiasing approaches is a promising direction for future work.

\newpage
\section{Samples from pretrained baseline models}

\begin{figure}[ht]
    \centering
    \includegraphics[width=0.93\textwidth]{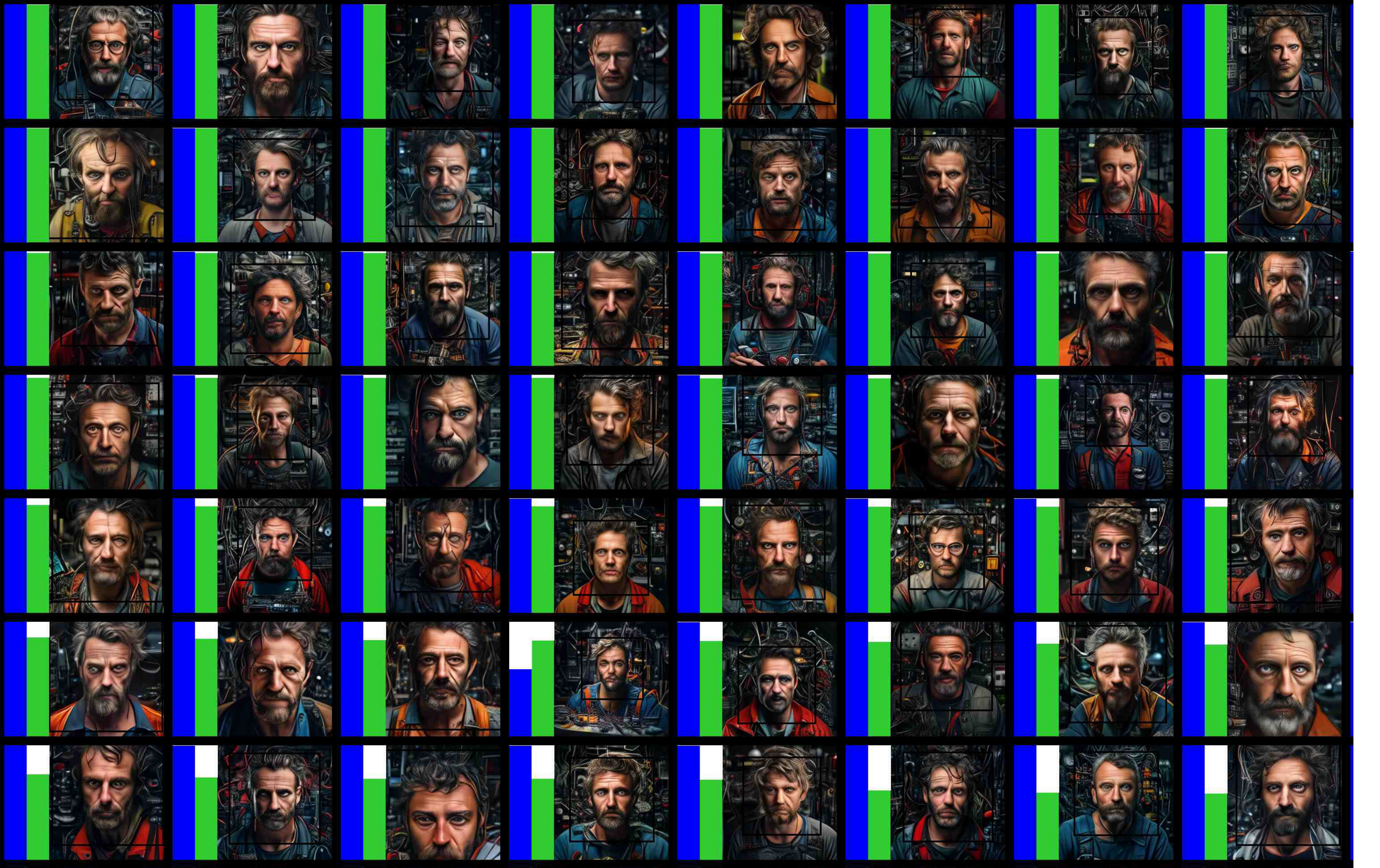}
    \caption{
    \looseness=-1
    \textbf{Given the test prompt: “a photo of the face of a electrical and electronics repairer, a person”, images generated by VILA-U.}
    \label{fig:vila-u}}
\end{figure}

\begin{figure}[ht]
    \centering
    \includegraphics[width=0.93\textwidth]{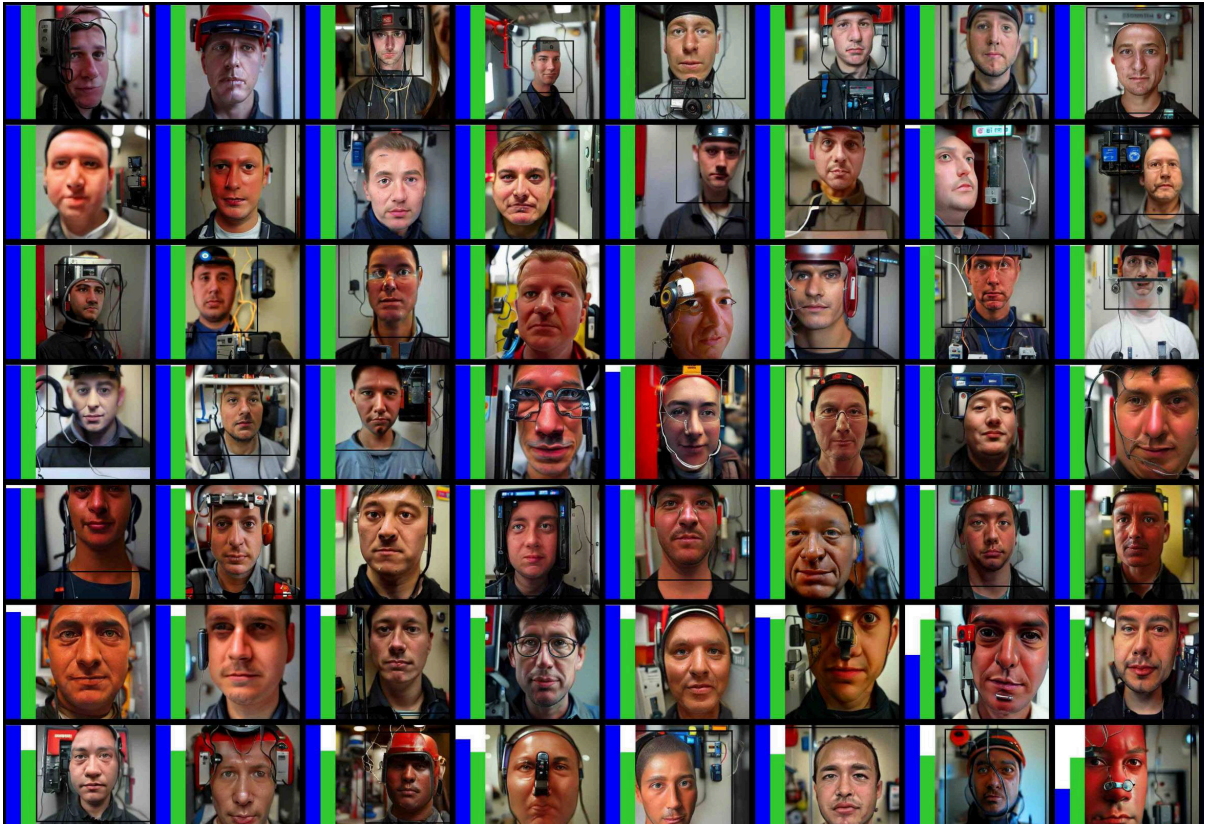}
    \caption{
    \looseness=-1
    \textbf{Given the test prompt: “a photo of the face of a electrical and electronics repairer, a person”, images generated by Janus}
    \label{fig:janus}}
\end{figure}

\begin{figure}[ht]
    \centering
    \includegraphics[width=0.93\textwidth]{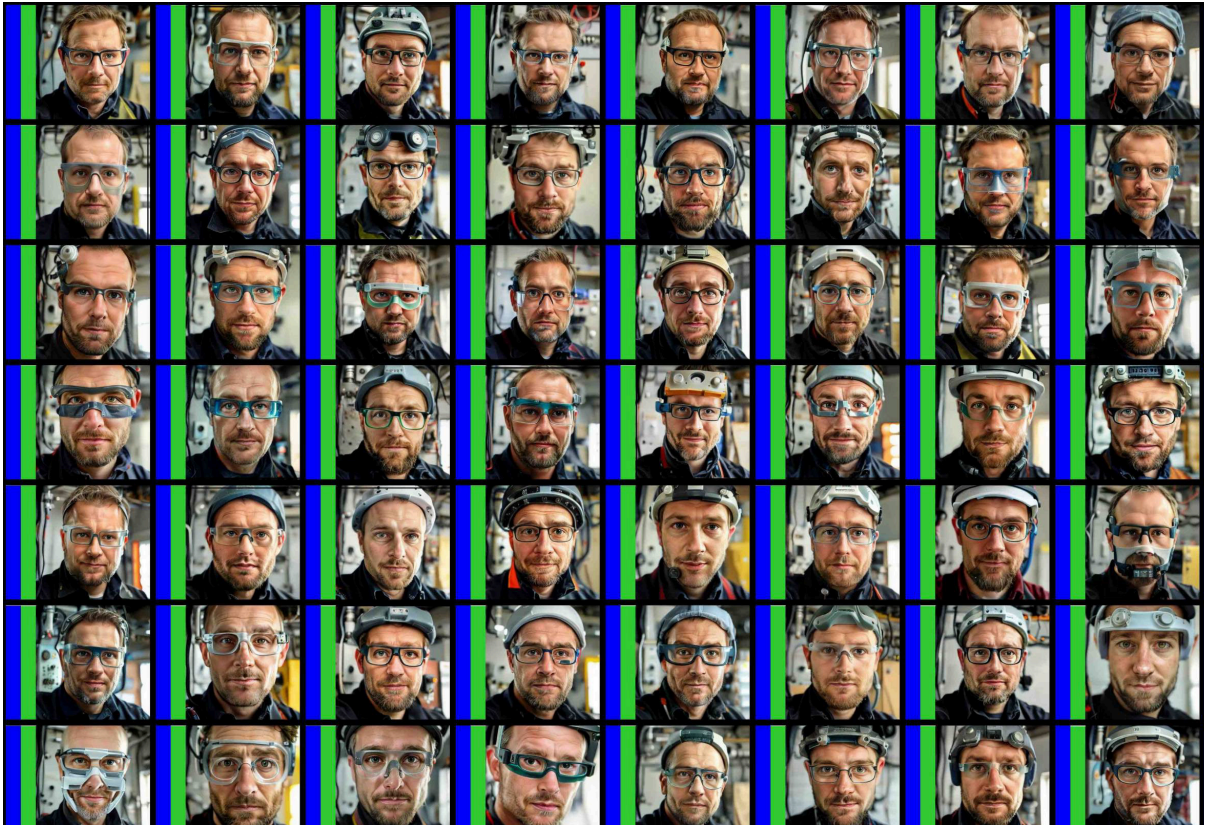}
    \caption{
    \looseness=-1
    \textbf{Given the test prompt: “a photo of the face of a electrical and electronics repairer, a person”, images generated by Janus-Pro}
    \label{fig:janus-pro}}
\end{figure}

\begin{figure}[ht]
    \centering
    \includegraphics[width=0.93\textwidth]{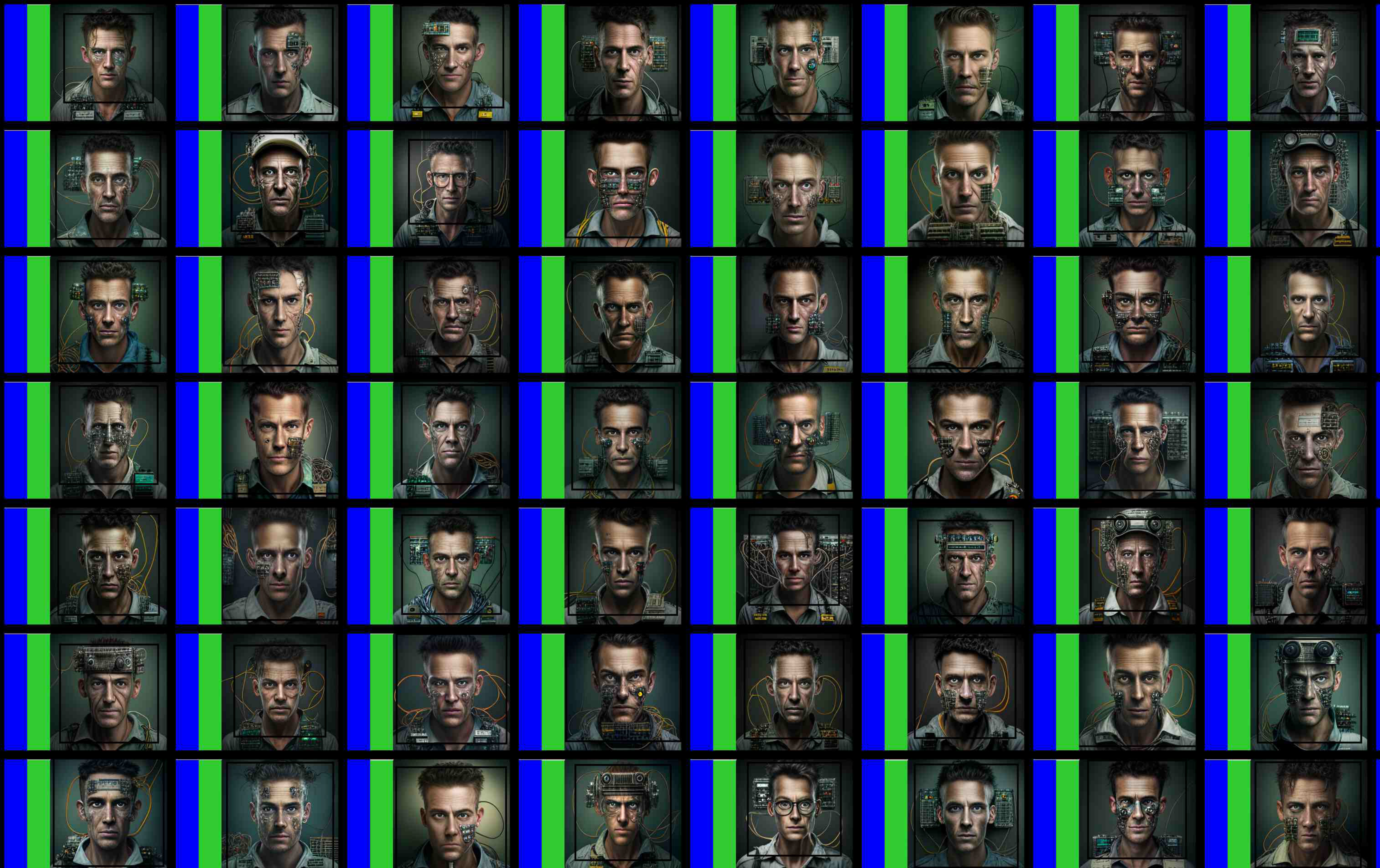}
    \caption{
    \looseness=-1
    \textbf{Given the test prompt: “a photo of the face of a electrical and electronics repairer, a person”, images generated by Show-o}
    \label{fig:show-o}}
\end{figure}

\begin{figure}[ht]
    \centering
    \includegraphics[width=0.93\textwidth]{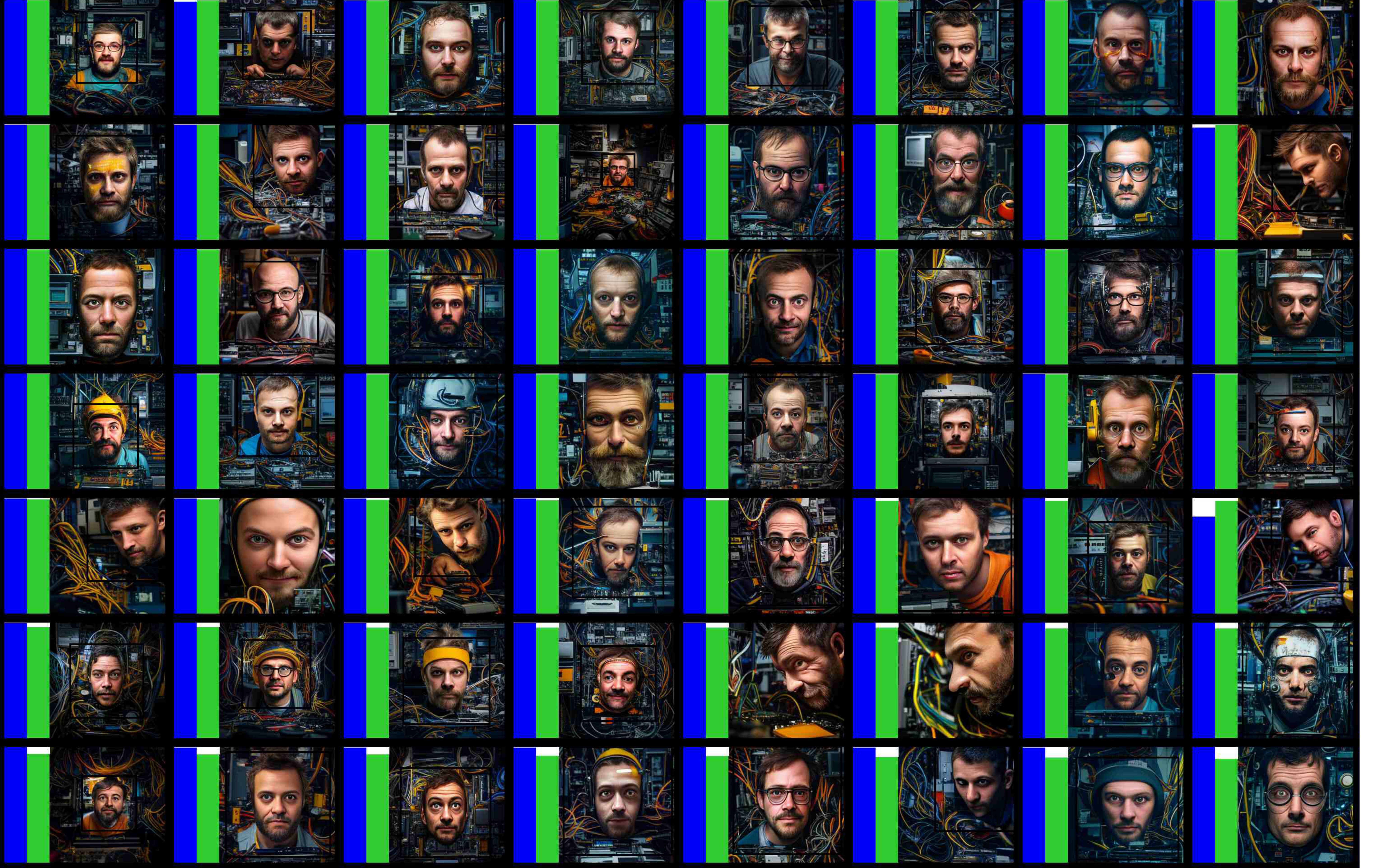}
    \caption{
    \looseness=-1
    \textbf{Given the test prompt: “a photo of the face of a electrical and electronics repairer, a person”, images generated by TokenFlow}
    \label{fig:tokenflow}}
\end{figure}

\begin{figure}[ht]
    \centering
    \includegraphics[width=0.93\textwidth]{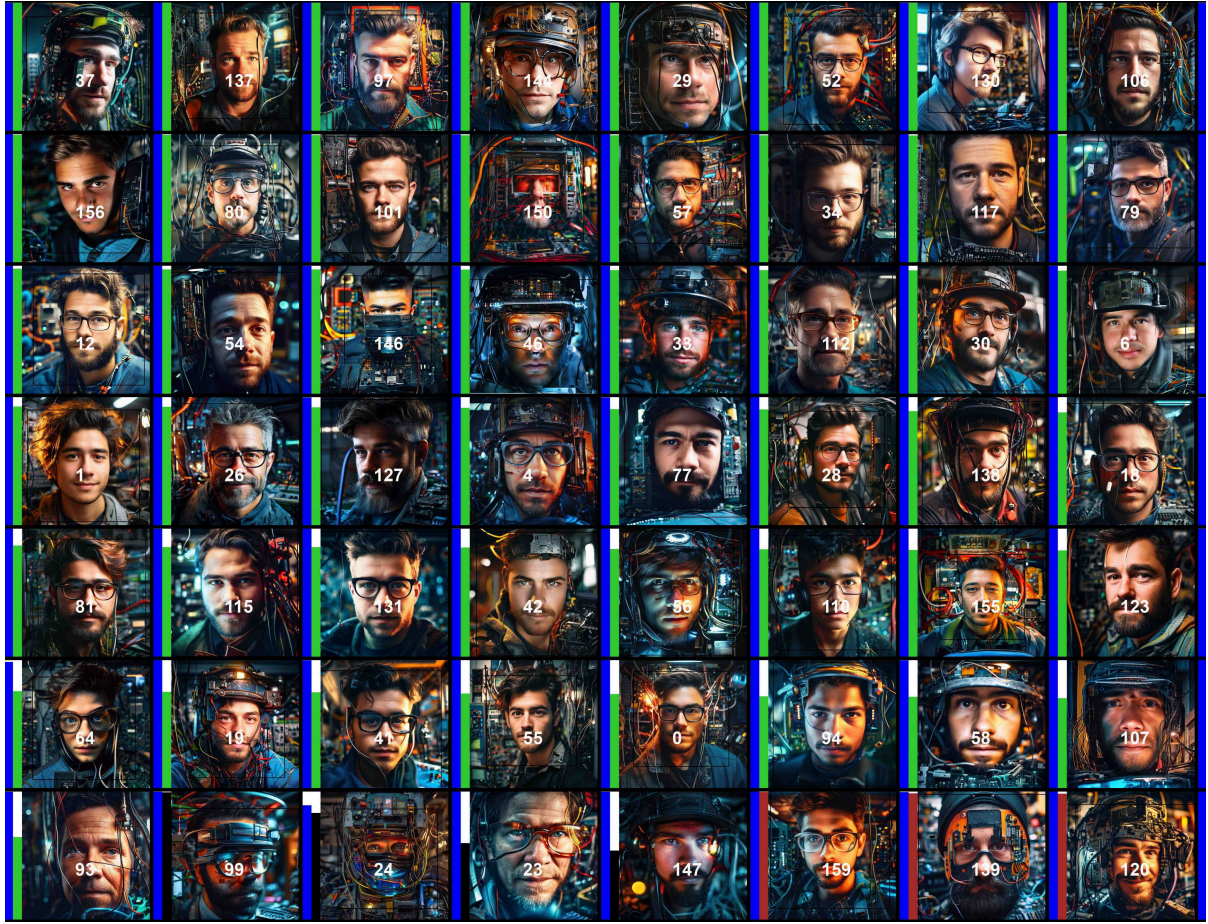}
    \caption{
    \looseness=-1
    \textbf{Given the test prompt: “a photo of the face of a electrical and electronics repairer, a person”, images generated by Emu3}
    \label{fig:emu3}}
\end{figure}


\end{document}